\documentclass[10pt, conference, compsocconf]{IEEEtran}
\usepackage{cite}
\usepackage{algorithmic}
\usepackage{graphicx}
\usepackage{textcomp}
\usepackage{xcolor}
\usepackage{microtype}
\usepackage{graphicx}
\usepackage{subfigure}
\usepackage{booktabs} 
\usepackage{url}
\usepackage[cmex10]{amsmath}
\DeclareMathOperator*{\argmax}{arg\,max}
\DeclareMathOperator*{\argmin}{arg\,min}

\ifCLASSINFOpdf
\else
\fi

\begin{document}
%
\title{Lifelong Learning Without a Task Oracle}


\author{\IEEEauthorblockN{Amanda Rios}
\IEEEauthorblockA{
\textit{University of Southern California}\\
Los Angeles, USA \\
amandari@usc.edu}
\and
\IEEEauthorblockN{Laurent Itti}
\IEEEauthorblockA{
\textit{University of Southern California}\\
Los Angeles, USA \\
itti@usc.edu}
}

\maketitle
\footnote{Proceedings of the IEEE 32nd International Conference on Tools with Artificial Intelligence (ICTAI 2020).}
\begin{abstract}
Supervised deep neural networks are known to undergo a sharp decline in the accuracy of older tasks when new tasks are learned, termed “catastrophic forgetting”. Many state-of-the-art solutions to continual learning rely on biasing and/or partitioning a model to accommodate successive tasks incrementally. However, these methods largely depend on the availability of a task-oracle to confer task identities to each test sample, without which the models are entirely unable to perform. To address this shortcoming, we propose and compare several candidate task-assigning mappers which require very little memory overhead: (1) Incremental unsupervised prototype assignment using either nearest means, Gaussian Mixture Models or fuzzy ART backbones; (2) Supervised incremental prototype assignment with fast fuzzy ARTMAP; (3) Shallow perceptron trained via a dynamic coreset. Our proposed model variants are trained either from pre-trained feature extractors or task-dependent feature embeddings of the main classifier network. We apply these pipeline variants to continual learning benchmarks, comprised of either sequences of several datasets or within one single dataset. Overall, these methods, despite their simplicity and compactness, perform very close to a ground truth oracle, especially in experiments of inter-dataset task assignment. Moreover, best-performing variants only impose an average cost of 1.7\% parameter memory increase.
\end{abstract}


\IEEEpeerreviewmaketitle

\section{Introduction}
Throughout life, humans are presented with unknowns in the environment, to which they must adapt and learn from. Yet, lifelong novelty integration must always coexist with strong mechanisms of protection against interference to consolidated learning, the stability-plasticity balance. Lifelong Learning remains a persistent challenge for artificial intelligence. State of the art deep neural networks are known to undergo a phenomenon termed “catastrophic forgetting”, which describes a drastic decline in the performance of the model on previously learned tasks as soon as a novel task is introduced \cite{french}. A straightforward reason is that if data used for acquiring previous knowledge is no longer present when assimilating a new task, gradient descent will optimize the model's weights under an objective that pertains only to the new samples. This may lead to a parameterization that grossly deviates from previous tasks' optimal states, and subsequent memory erasure.

In recent literature, several methods have been proposed to mitigate catastrophic forgetting, many of which rely on partitioning or biasing a network to accommodate successive tasks incrementally \cite{cheung,wen,zeng, mallya}. However, these approaches require an oracle to assign task identities to incoming samples and re-tune the model to an appropriate task-dependent response. In fact, we show that if an oracle is removed from a task-dependent model, performance starkly declines. Such a dependency is a latent impediment towards more realistic lifelong learning settings where task assignments would ideally be inferred in an end-to-end manner or would not be required at all.

Methods for substituting a task oracle have not been extensively studied or discussed in the field. Incipient proposals and discussions have occurred in \cite{gepperth, oswald, aljundi2017} but in general  their scalability to difficult benchmarks is poor. 
To fill this gap, here we propose and compare several models that impose only a very restricted parameter memory increment with respect to the base task-dependent fine-grained classification model. Moreover, we do so in an equalized context, using the same architecture backbones among task mappers and task-independent baselines, while weighing memory-performance trade-offs of each approach. Lastly, emphasizing that task mapping is itself subject to catastrophic forgetting, we show that task assignment can have largely varying degrees of difficulty depending on the incremental learning paradigm used. For this we distinguish between two overarching domains of incremental learning protocols, (1) learning categories within one dataset versus (2) learning categories over different datasets. 

Overall, we find that when using our best performing task mapper coupled with a state-of-the-art fine-grained classifier, we can perform better than current  task-independent methods tested and at much lower relative memory expenditure. 

\section{Background}

We focus on a continual learning paradigm where a single neural network must incrementally learn from a series of tasks and, each time a new task is learned, access to previous tasks’ data is limited or absent. We assume that all tasks are unique and clearly separated. Within this framework, we can interpret most of the recent continual learning (CL) literature as being grouped into models which rely on information of task identity at test time, termed “task-dependent”, and those which do not, “task-independent” (Figure 1).

\subsection{Task-Independent Continual Learning}

This class of algorithms does not require task labels at test time and typically operates on a single-head architecture, i.e., the final layer has as many nodes as classes over all tasks and each task trains on shifted labels of its original classes. 
Nonetheless, task-independent models can also operate with a single shared output head \cite{ven}. 

Within task-independent methods a major subclass are regularization approaches, which constrain the change of learnable parameters to prevent "overwriting" what was previously encoded. For instance, Elastic Weight Consolidation (EWC) \cite{kirkpatrick} computes a Fisher information matrix at each task switch to penalize changes to highly correlated weights. In a similar vein, Synaptic Intelligence \cite{zenke} uses path integrals of loss derivatives and a minimum combined loss to constrain crucial parameters.
Another important subclass are replay algorithms. These methods purposely approximate the continual learning framework to a multi-task setting by estimating  past data distribution either via storage of a select memory coreset \cite{lopez, rebuffi} or by reconstruction via training a generative model to sample past data and labels \cite{shin,rios}. 

Overall, because task-independent models do not make use of task-specific parameters at test time, the classification endeavour is naturally more challenging. For example, regularization approaches such as EWC have been shown to perform poorly in single-headed scenarios \cite{parisi}. Replay approaches can quickly escalate to expensive memory usage. Generative replay methods are also limited by the output quality of the data generators themselves, which are, furthermore, subject to forgetting. Finally, task-independent algorithms may still be converted to task-dependent if task-specific elements are added. An example is using regularization in the main network but adding independent output heads. In such cases, even though the algorithms themselves do not require task-IDs, the added task-specific elements will.

\begin{figure}[t]
    \begin{center}
        \includegraphics[width=7.7cm]{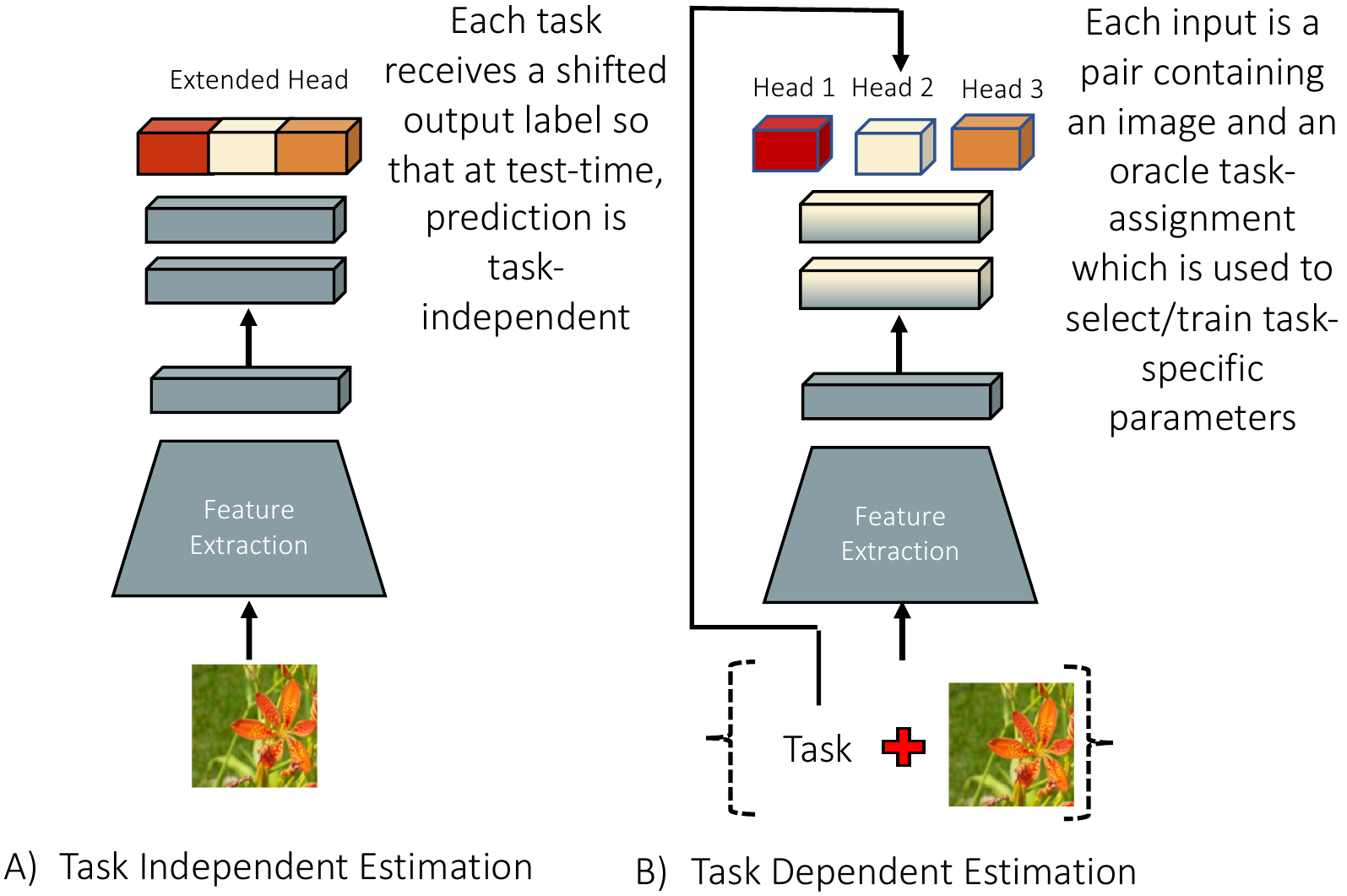}
		\caption{\textbf{(A) Task-Independent (B) Task-Dependent models.} We show the most common architectural schemes for the two modalities. Task-dependent models require task labels as input. Task-independent do not.
		}
        \label{fig:model}
    \end{center}
    \vspace{-15pt}
\end{figure}

\subsection{Task-Dependent Continual Learning}

Task-Dependent algorithms rely on oracle-generated task labels both for training and testing. Because of this, models can explore a range of task-specific components. For instance, most task-dependent models include one separate output head per task and employ various mechanisms to share the remainder of the network. A common strategy is to make use of task-dependent context \cite{zeng, cheung} or mask matrices \cite{mallya} to partition the original network for each task. In general, partitioning methods come at the cost of large storage overheads \cite{zeng}.

To counteract memory limitations, Cheung et al \cite{cheung} propose diagonal  binary context matrices and perform parameter superposition (PSP) at very low memory cost. Finally, Wen et al \cite{wen} attempt to minimize performance decay over cumulative tasks by computing task-specific layer-wise beneficial directions (BD) that bias the network towards correct classification. BD's are inspired by an inverse of the commonly known adversarial directions, in this case to force apart classification boundaries which might otherwise become tangled during incremental learning. BD is not meant as a standalone method, but provides significant boost in performance when paired with other models such as PSP.

\section{Methods - Learning without a Task Oracle}

The major limitation of task-dependent methods is their inability to perform without oracle input. In fact, determining task ID's is a learning procedure itself subject to catastrophic forgetting. We propose several task-mapper algorithms to substitute the oracle input required for task-dependent continual learning. Our models prioritize simplicity and low memory usage insofar as they are meant as an add-on to a base fine-grained classifier. However, we show that our best mappers are sufficient in obtaining very good task estimation accuracies.

To evaluate our task mappers and  baselines, we use parameter superposition (PSP - \cite{cheung})  with  Beneficial  Biases  (BD  -  \cite{wen}) as a standard state-of-the-art task-dependent backbone for fine-grained classification. At  test  time,  the  task mapper predicts a task label which activates task-specific PSP context keys and BD biases (Fig. 2). Both our fine-grained classifier and our task mappers receive as input a shared fixed-feature representation from a CNN (Resnet or other) pretrained on ImageNet. One reason is that, biologically, low-level visual features are thought to be optimized during evolution and early development and later re-utilized, i.e., are task nonspecific and do not need to be constantly re-learned. In contrast, higher-level visual features are often task-specific and build upon a recombination of early-level input \cite{petrov}. In fact, many recent works on meta and adaptive learning have shown state-of-the-art performance when re-utilizing a fixed common feature embedding for all novel tasks \cite{dhillon}. Figure 3 shows our proposed methods.

\begin{figure}[t]
    \begin{center}
        \includegraphics[width=5.1cm]{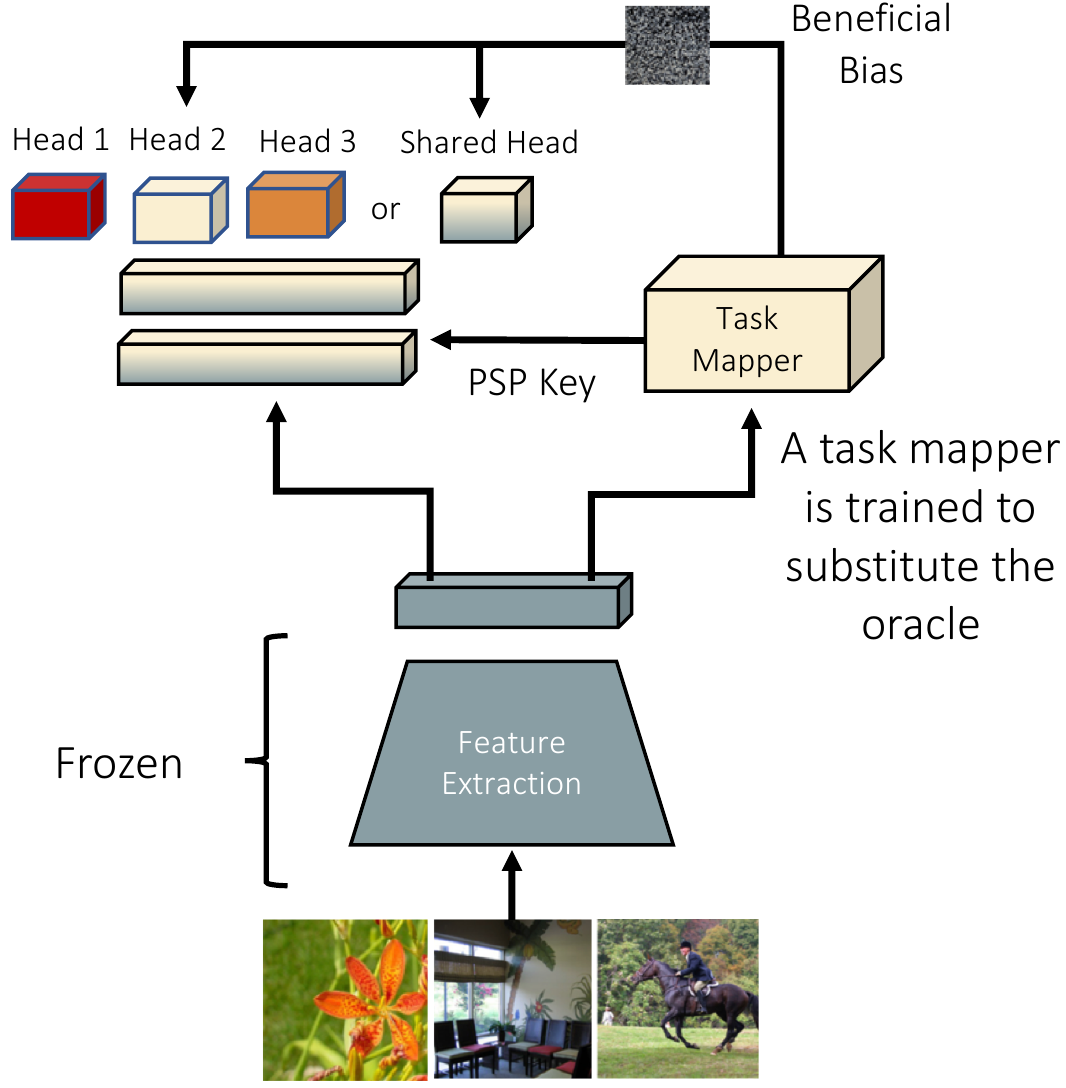}
		\caption{\textbf{Outline of our pipeline.} We employ parameter superposition (PSP- Cheung et al, 2019) with Beneficial Biases (BD - Wen et al, 2020) as a state-of-the-art task-dependent backbone for fine-grained classification. We then substitute the oracle input for an end-to-end trainable task-mapper. We experiment with different low-memory task-mapper variants. In all the variants, at test time, the task-mapper predicts task assignments which are then used to activate task-specific PSP context keys and BD biases.}
        \label{fig:model}
    \end{center}
    \vspace{-15pt}
\end{figure}

\begin{figure*}[t]
    \begin{center}
        \includegraphics[width=17.7cm]{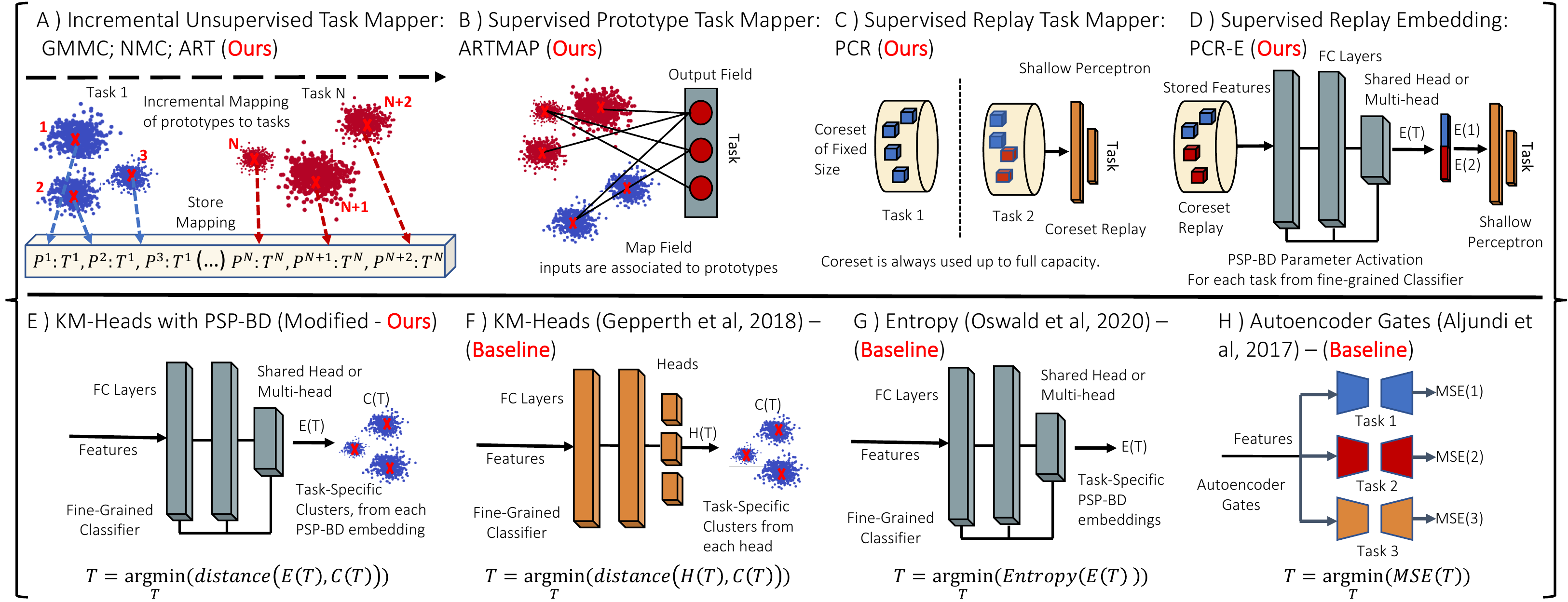}
		\caption{
		\textbf{Task-Mapping Models}. Mappers receive features taken from a feature extractor network that is shared with the pipeline's fine-grained classifier. A) Incremental versions of ART, Gaussian Mixture Classifier (GMMC) and Nearest  Means (NMC) - prototype-based networks which all employ a dictionary mapping of prototypes to task labels. B) Fast ARTMAP architecture trained to form feature to task mapping directly. C) (PCR): maps features to task by a one layer perceptron with aid of a very small coreset for replay. D)(PCR-E) - we sift through  PSP-BD  keys  and  biases to  collect  all  task-specific  output  responses  which  are  then concatenated and used as input to a shallow perceptron. Features are stored in a coreset similar to (C). E) (KM-heads Ours) is a proposed modification to the baseline model in Gepperth et al - we sift through PSP-BD task activated embeddings to perform task-conditioned K-means and the task is assigned as the closest prototype's task label. F) (KM-heads, Gepperth et al) is similar to (E) except, there are no PSP-BD elements and output is necessarily multi-head. G) (Entropy, oswald et al, 2020) - task assignment is given by the index of the PSP-BD embedding that yields lowest predictive uncertainty. H) (AE-Gates, Aljundi et al) Separate Autoencoders (AEs) are trained for each task. During testing, a sample image is reconstructed by each of the task-dependent AEs and the final task assignment is given by the AE which yielded the smallest mean squared reconstruction error (MSE).
		}
        \label{fig:model}
    \end{center}
    \vspace{-5pt}
\end{figure*}

\subsection{Incremental Unsupervised Task Mappers}

Different unsupervised prototype-based networks are employed to cluster pre-extracted features from the current task. Prototype-based networks have as an advantage that they encode information much more locally than conventional deep networks (DNs), i.e., in each individual prototype. DNs rely on a highly distributed mapping between thousands of network weights. 
In incremental learning, localization naturally minimizes inter-task disruption. In fact, for single-headed continual learning, most approaches employing DNs have been shown to require a minimum degree of data replay to prevent overwriting \cite{parisi}. Nonetheless, prototype networks underperform DNs in classic supervised classification. Thus, we leverage prototype-based networks only for incremental task-mapping, a form of coarse-level identification. By coupling them to an efficient task-dependent DN fine-grained classifier, we seek a complementarity that enables more efficient lifelong learning. 

\subsubsection{Nearest Means Classifier (NMC)}

In our simplest prototype-based approach, to learn a new task we start with a fixed embedding of this task and perform K-means clustering. The K resulting prototypes all receive an attached super-label equal to the current task's identity. As new tasks pile up, to perform task prediction at any given time, we keep a running dictionary, $D_{map}$, of the cluster to task mapping:
\begin{equation}
D_{map}(t) = \left\{(m^{i}:1),...,(m^{K}:1),...,(m^{Kt}:t)\right\}
\end{equation}

Since feature representation is fixed, this mapping will not change over time. For any given sample, we find the closest stored prototype and use its task label as the predicted task:
 
\begin{equation}
Task = D_{map}(\min_{i}(||m^{i} - x^i||^2_2))
\end{equation}

\subsubsection{Gaussian Mixture Model Classifier (GMMC)}

Here we employ Gaussian Mixtures (GMs) to perform task-wise incremental prototype generation. The advantages of using GMs over nearest means is that they additionally encode variance information, and, furthermore, perform soft-assignment, which renders them more robust to outliers during clustering. The incremental task-mapping algorithm is very similar to NMC, differing only in how the prototypes are generated. For each new task, K GM prototypes are computed from the extracted fixed embedding of that task, with overall distribution:
\begin{equation}
f_t(x;\mu,\Sigma) = \sum_{i=1}^{K}w_i\cdot N(x;\mu_{i},\Sigma_{i})
\end{equation}
where $\mu_i, \Sigma_i$ are mean and covariance of each of the K Gaussian distributions. We also save K Gaussian weights, $w_i$, per task. At each task switch we re-normalize them to:
\begin{equation}
\frac{w_i\cdot K}{\sum_{i=1}^{KT}(w_i\cdot K)} \quad \forall i
\end{equation}

The task mapper then becomes a dictionary containing assignments from GM parameters to task super-labels:
\begin{equation}
D_{map}(t) = \left\{(N_{i},w_i:1),...,(N_{kT},w_{kT}:t)\right\}
\end{equation}
where $N_i$ refers to one of the Gaussians in the accumulated task-mapping function and $w_{i}$ to its mixture weight out of $K\cdot T$ weights. Task prediction occurs similarly as before:
\begin{equation}
Task = D_{map}(\argmin_k(P(k,x_i)))
\end{equation}
where the probability of sample $x_i$ belonging to the $k^{th}$ gaussian is given by:
\begin{equation}
P(k,x_i) = \frac{w_k N(x;\mu_{k},\Sigma_{k})}{\sum_{n=1}^{KT}w_n N(x;\mu_{n},\Sigma_{n})}
\end{equation}

\subsubsection{Fuzzy ART Classifier}

In this variant we generate the incremental prototypes with an unsupervised fuzzy ART network \cite{carpenter,vakil}. ART networks were initially proposed to overcome the stability-plasticity dilemma by accepting and adapting a stored prototype only when an input is sufficiently similar to it. In ART, when an input pattern is not sufficiently close to any existing prototype, a new node is created with that input as a prototype template. Similarity depends on a vigilance parameter $\rho$, with $0 < \rho <1$. When $\rho$ is small, the similarity condition is easier to achieve, resulting in a coarse categorization with few prototypes. A $\rho$ close to 1 results in many finely divided categories at the cost of larger memory consumption. One further specification of ART is that an input $x$ of dimension $D$ undergoes a pre-processing step called complement coding, which doubles its dimension to $2D$ while keeping a constant norm, $x^* = [x,\vec{1}-x]$. This procedure prevents category proliferation \cite{carpenter} but makes each prototype also occupy double amount of space. 

During learning of each task, if for $x$ a prototype $w_i$ is sufficiently similar by satisfying:
\begin{equation}
\frac{||\min(x,w_i)||_1}{||x||_1} > \rho 
\end{equation}
then $w_i$ can be updated according to:
\begin{equation}
w_i(t+1) = (1-\beta)w_t(t) + \beta (\min(x,w_i(t)))
\end{equation}

To adapt an unsupervised ART for incremental task classification, as a new task is learned, we set $\rho=1$ for all prototypes of the already learned tasks. By doing this we allow updates only to the current tasks' freshly created prototypes, shielding previous tasks' information from interference. Similarly to sections A-1 and A-2, we also keep a running dictionary, $D_{map}$ with task mappings between prototypes and their corresponding task super-labels. Task is predicted as: 
\begin{equation}
Task = D_{map}(\argmax_i(\frac{||\min(x,w_i)||_1}{\alpha+||w_i||_1}))
\end{equation}
where $\alpha$ is a regularization hyperparameter that penalizes larger weights. We found $\alpha=0.001$ worked best.

\subsection{Supervised Prototype Mapping (ARTMAP)}

This variant is a natural extension to our unsupervised ART model, which employs a supervised fuzzy ART (ARTMAP) architecture \cite{vakil}. The advantage of using an ARTMAP for task mapping is it they naturally allows for incremental learning without interference to previous prototypes. Whereas in ART we imposed a prototype-specific $\rho$ parameter freezing, in ARTMAP this is no longer necessary since updates only occur if a prototype has the same label as the training sample. Otherwise, in case of a category mismatch, the vigilance is adjusted temporarily, called match tracking:

\begin{equation}
\rho_{temp} = \frac{||\min(x,w_i)||_1}{||x||_1} + \epsilon
\end{equation}

This can be repeated as many times as necessary until finding a label match or until $\rho_{temp}=1$, at which point a new sample $x$ is drawn, the current one discarded, and $\rho_{temp} = \rho$. At test time, the task predicted is the label of the closest prototype.  

\subsection{Perceptron with Coreset Replay (PCR)}

In this model, we incrementally map feature arrays to task assignments using a shallow perceptron aided by replay from a fixed-size memory coreset. The features used as inputs are obtained via transfer-learning from a frozen feature extractor. The pre-trained feature extractors were selected to generate feature arrays with a much lower dimensionality then the original input images, reducing the memory burden on coreset size as well as network size. 

We hypothesized that using coreset replay only to perform coarse level classification, i.e., task-mapping, instead of fine-grained incremental classification, could radically cut costs of memory storage overall. The intuition is that covering coarse-level data statistics inside a small coreset can be comparably a much easier endeavour. Thus, in a full pipeline,
we could obtain a better performance-to-memory ratio overall.  

\subsubsection{Coreset Building}
At the time of insertion into the memory buffer, we select a number of feature vectors from the new task equal to N/T(t), where N is size of coreset and T(t) is total number of tasks learned until then. Additionally, since the coreset is always filled to full capacity, at each task-switch we re-compute the per-task allowance and remove an equivalent number of old feature vectors per task, maintaining a homogeneous task representation in the coreset at all times. We experimented with other coreset building techniques such as homogeneous coreset sampling according to center means and ART prototypes, but they did not perform as well (additional results in supplementary - \cite{supplementary}). We hypothesize that this was because the coresets used were very small, and it was more important to guarantee homogeneous task sampling than feature-level prototype diversity.
\subsubsection{Perceptron Training}
At each task, our model is trained using an extended dataset which includes feature vectors from the new task and those from past tasks that are contained in the memory coreset, forming an extended training set. The final loss is optimized by stochastic gradient descent: 
\begin{equation}
Loss_t = {Loss_{t}^{new} \cup \lambda_{mem} Loss_{t-1}^{memory}}\label{eq}
\end{equation}
where $\lambda_{mem}$ weighs the importance of old tasks relative to new. We found that dynamically setting $\lambda_{mem} = \frac{T_{memory}}{T_{all}}$ during training worked best. $T_{memory}$ refers to the number of tasks in memory and $T_{all} = T_{memory} + T_{new}$. We train in single-head mode: one output node per task. During training, the perceptron is not reinitialized, to enable forward transfer of knowledge. For architecture, we found that a 1-layer perceptron outperformed memory-equivalent deeper perceptrons (supplementary - \cite{supplementary}).

\subsection{Perceptron + replay of task-specific embeddings (PCR-E)}

Previously proposed methods \cite{oswald, gepperth} use the final layer task-specific embedding as basis for a task-mapping heuristic. Their hypothesis is that the last-layer activity space for the correct task will differentiate itself statistically from the others. In their case, this is measured via entropy or prototype distances, but here we also propose to map  last-layer task-specific embeddings to task predictions via a shallow perceptron. As such, we use a PSP-BD fine-grained classifier and sift through PSP-BD keys and biases to collect all task-specific output responses which are then concatenated and used as input to a shallow perceptron (one hidden layer). At each task, the perceptron learns to map this concatenated array to the correct task label. To not forget previous task mappings, we keep a coreset with previous tasks' feature arrays as previously described in section C-1. These samples are processed in the same way as new inputs, by also sifting through task-specific keys and collecting the concatenated output responses. 

\begin{figure*}[t]
    \begin{center}
        \includegraphics[width=16cm]{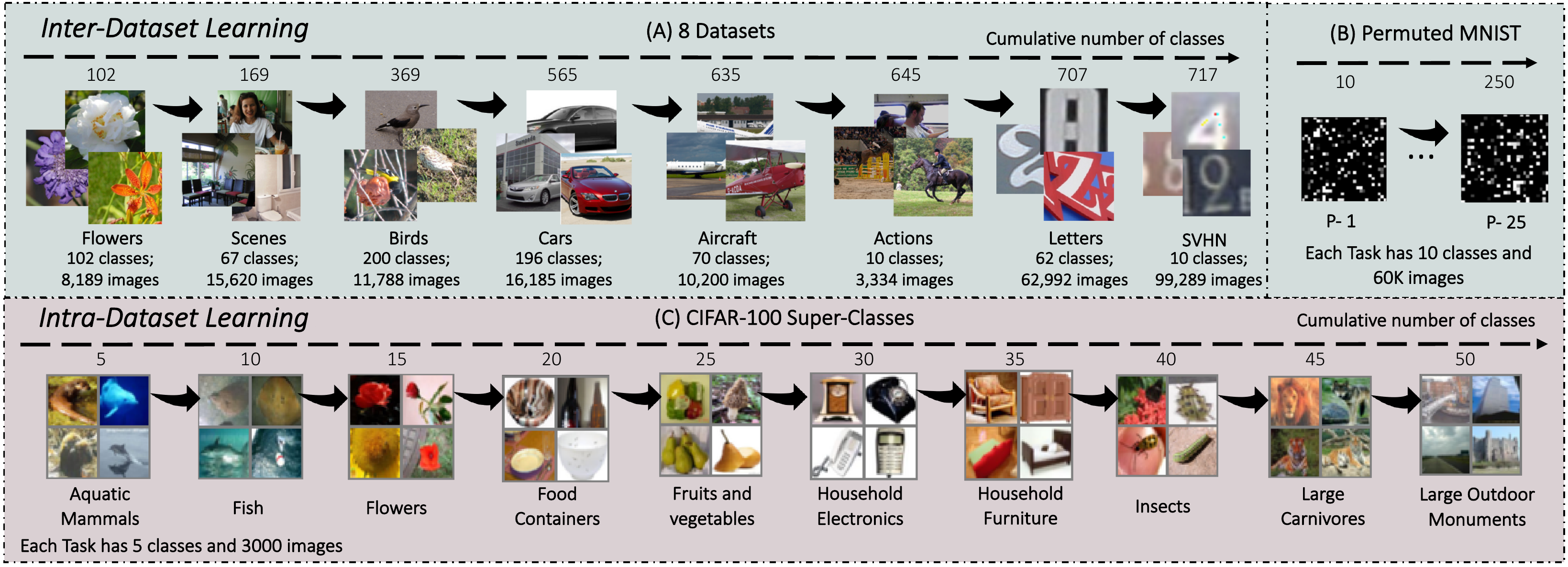}
		\caption{
		Schematic of our incremental learning experiments, which can be divided into two main categories, Inter-Dataset and Intra-dataset. For Inter-dataset, each task is a complete dataset: (A) A sequence of 8 datasets containing natural images \cite{aljundi}. (B) A sequence of 25 datasets, each a permutations of MNIST. In the Intra-Dataset modality, each task is a subset of one dataset. Here we use 10 different super-classes of Cifar-100 each as a separate task.  }
        \label{fig:model}
    \end{center}
    \vspace{-5pt}
\end{figure*}

\subsection{Task-Mapper Baselines and Baseline-Modifications}



Few works in the literature have addressed task mapping, and their coverage of more difficult benchmarks is very limited.
We compare our approaches to three recently proposed methods [Entropy - \cite{oswald}; Autoencoder Gates - \cite{aljundi2017}; Multi-Head KM - \cite{gepperth}]. We also include upper and lower bound baselines which consist of training the task-dependent fine-grained classifier with ground truth task labels and random task assignments, respectively. Finally, we set as an additional baseline a modified version of Multi-Head KM which includes other relevant task-specific elements such as PSP partitioning and BD biases. For all baselines we always employ the same fine-grained classifier architecture and optimizer. The only differences appear in the layout of the output heads which depend on the particulars of each proposed method.

\subsubsection{Baseline - Entropy}

This algorithm is adapted from \cite{oswald} who use an output layer's predictive uncertainty to establish task identity. 
In our implementation, given an input pattern, we sift through PSP task partitions and task-specific BD biases for all tasks seen so far, obtaining multiple task-conditioned embeddings. We predict task assignment by the index of the task embedding, $f(x,\theta_t)$, which yields lowest predictive uncertainty, i.e., lowest output entropy: 
\begin{equation}
{task = \argmin_{t}(Entropy(f(x,\theta_t)))}\label{eq}
\end{equation}

\subsubsection{Baseline - Expert Autoencoder Gates (AE-gates)}
We adapt the algorithm proposed in \cite{aljundi2017}. Our implementation details and parameters can be found in \cite{supplementary}. In this model, one single-layer undercomplete autoencoder (AE) is trained separately for each task, capturing shared task statistics in it's latent space encoding. 
At test time, we pass a test sample through all AE's meanwhile computing for each AE the mean squared reconstruction error (MSE). The final task prediction is given by the AE with lowest MSE:

\begin{equation}
{task = \argmin_{t}(MSE(AE_t(x)))}\label{eq}
\end{equation}

\subsubsection{Baseline - Clustering in Multi-Head Outputs (KM-heads)}

We adapt the algorithm in \cite{gepperth}. In this model, one separate output head is created for each task, for a total of $T$ heads. After supervised training of current task $t$, a forward pass with the current tasks's data gives an embedding of head $H(t)$ which is then clustered via k-means. The resulting prototypes are stored. This procedure is repeated after each task is learned, resulting, at time $t$, in a number $T(t)$ of different embeddings as well as $T(t)\cdot N$ prototypes. At test time, task is predicted by running a sample through the network and, at each head, computing the minimum distance from that sample to the closest head-specific prototype. The overall task prediction is given by the absolute minimum distance from all heads. The winner head is then used to perform fine-grained classification:

 \begin{equation}
{task = \argmin_{t}[\min_{n}(\lVert (H_{t}(x) - P(n))\rVert )]}\label{eq}
\end{equation}

\subsubsection{Baseline Modification - Clustering in Multi-Head or Shared-Head with PSP-BD Task-Partitioning (KM-heads Ours)}

In KM-heads the only task-specific elements are heads and the remainder of the network is shared. Here we modify the algorithm by adding other task-specific elements such as PSP partitioning and BD biases. By adding these components we enable two versions of readout, one with multiple heads and the other with a shared head for all tasks.

\vspace{-2pt}
\subsection{Task Independent Baselines}
We evaluate how our pipeline (task mapper + PSP-BD fine-grained classifier) performs compared to  task-independent methods. We implement EWC - \cite{kirkpatrick}) and GEM – \cite{lopez} in a completely task-independent manner. Additionally, in reference to our coreset-replay task-mapper of section C, we propose task-independent vanilla replay in which we create a coreset as in C-1, but populate it homogeneously among classes. During training, coreset samples are interleaved with the new classes. For all baselines, we test versions where the output layer contains one node per class versus having a shared head with as many nodes as number of classes of the most populous task.

\section{Experiment Descriptions - Datasets}
Our experiments are divided into two categories: Inter and Intra-dataset. In Inter-dataset, each task is a complete dataset whereas in Intra-Dataset, a task is a subset of one dataset. 

\vspace{-3pt}
\subsection{8 Datasets experiment (Inter-Dataset)} We consider a sequence of eight object recognition datasets (Figure 4), with a total of 227,597 pictures in 717 classes as in \cite{aljundi}.
In this experiment, the frozen feature extractor corresponds to convolutional layers of Alexnet pretrained on Imagenet. For fine-grained classification, we use the complete embedding which is a 256x6x6 feature array, but, for task-mapping, we perform complete spatial pooling to only 256 units. This pooling operation did not significantly impact performance but greatly reduced storage requirements. For classification, we use the two last fully-connected layers of Alexnet, 4096 units each. We include one separate output layer per task as in \cite{aljundi}.

\vspace{-2pt}
\subsection{Permuted MNIST (Inter-Dataset)}
This experiment is formed by 25 datasets generated from randomly permuted handwritten MNIST digits. Each new task (dataset) has 10 classes. Since the datasets here are simple enough, our task mappers are fed permuted images directly, without previous feature extraction. We use two fully connected layers of 256 as a PSP fine-grained classifier. Tasks share a head with 10 output nodes.

\vspace{-2pt}
\subsection{Sequence of 10 Cifar100 superclasses (Intra-Dataset)}
In this experiment, a task is equivalent to one super-class from Cifar100. We use a total of 10 tasks. Each super-class contains 5 different sub-classes that are semantically linked, i.e., different fish species are sub-classes of the fish super-class. Assigning a task label thus constitutes a high-level categorization of sub-classes. For the frozen feature extraction we use resnet-34 up to the penultimate layer, pre-trained on Imagenet. Resnet embedding size is 512. For fine-grained classification, we use a MLP with two hidden layers of 7680 and 4096 units, followed by a shared head of 5 output nodes.

\begin{figure}[t]
    \begin{center}
        \includegraphics[width=6.2cm]{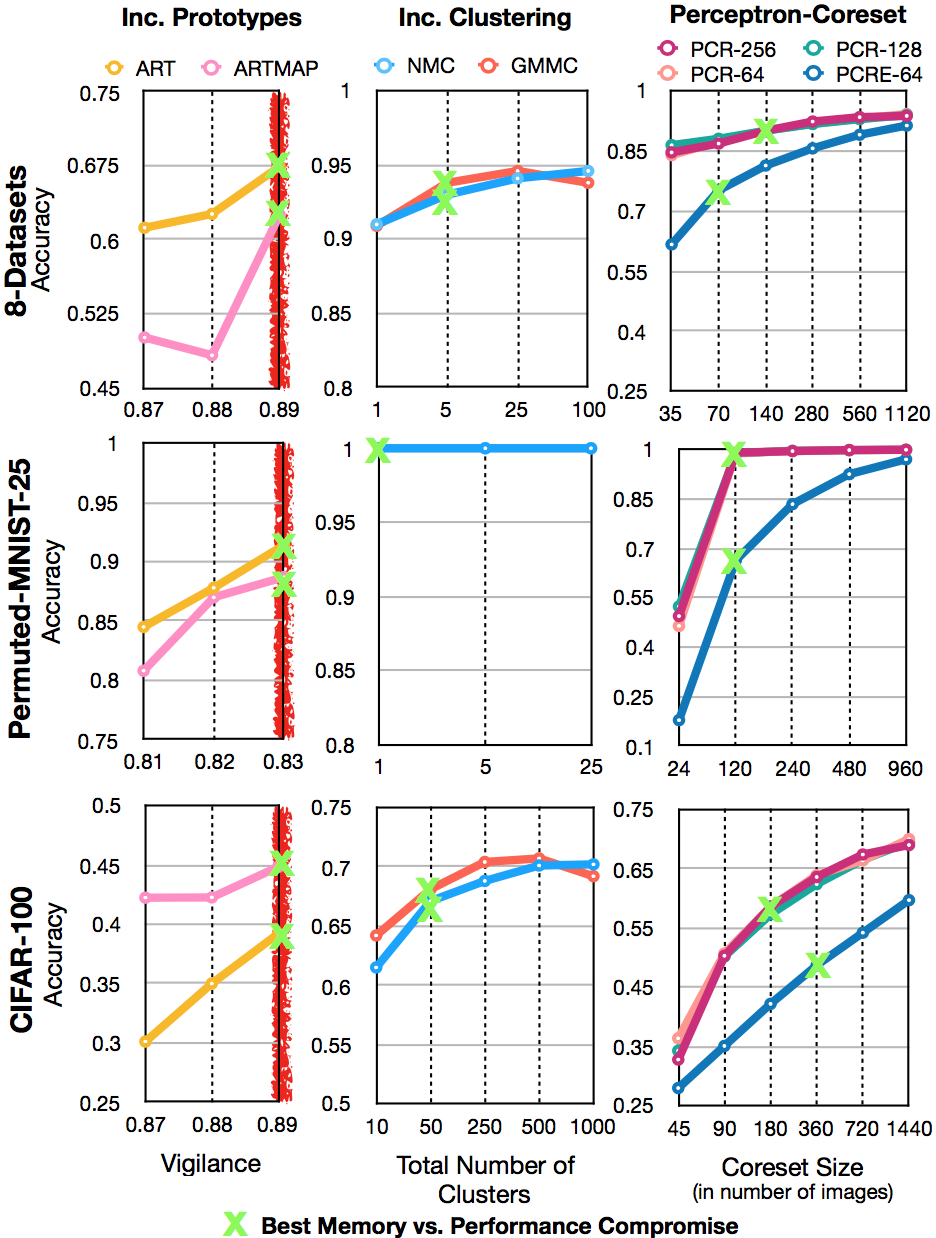}
		\caption{\textbf{Parameter Dependencies of the different task-mappers.} The vertical axes (task accuracy), have been scaled differently between each subplot so that variations to critical parameters are amplified. The green markers denote the best memory-performance tradeoff according to $score = A - \alpha M$. Most mappers achieve optimum memory-performance tradeoffs on an elbow point. However, for the first column (ART/ARTMAP), the vertical red line is a virtual memory wall, meaning we did not increase vigilance further because memory usage was already up to 10-fold larger than for all other task-mappers. Overall best memory-performance tradeoff is obtained by inc-GMMC and inc-NMC.}
        \label{fig:model}
    \end{center}
    \vspace{-10pt}
\end{figure}

\section{Results}
\subsection{Task Estimation - Parameter Dependency}
We analyze our task-mappers's performance-parameter dependency in figure 5, specifically for inc-GMMC, inc-NMC, PCR, PCR-E, inc-ART and ARTMAP.  For inc-ART/ARTMAP the actual number of prototypes per task is not determined {\em a priori} and varies according to the natural variance of each task. Instead, we vary vigilance parameter $\rho$, which controls how many prototypes are formed. The closest $\rho$ is to 1, the more prototypes are generated, reaching an upper bound where prototypes become copies of inputs. For GMMC and NMC we explicitly vary the number of prototypes per task. In the case of PCR and PCR-E we vary coreset size and the number of units in the hidden layer.

We select the best parameter configuration for each task mapper according to the memory-performance tradeoff as measured by: $score = A - \alpha M$, where A stands for task classification accuracy, M for  memory storage usage and $\alpha$ is the weight of memory usage, which was set to $10^{-6}$. We define memory usage as bytes required to store parameters and any appended data which will be used by the task mapper at all times (permanent). This is different then transient RAM usage during training. Detailed memory calculations in \cite{supplementary}.  

GMMC and NMC provided the best memory-performance tradeoff overall. In the 8-dsets and Cifar100 experiments, these methods used only 5 prototypes. For 8-dsets, GMMC occupied 82 Kilobytes while achieving 93.9\% task accuracy. For Permuted, the best mapper is an NMC with merely one prototype per task and virtually 100\% task-determination accuracy. Other methods such as inc-ART and ARTMAP show a poor absolute memory-performance tradeoff, where more prototypes naturally lead to better performance, but come at a huge memory cost. For example, with  $\rho=0.9$, inc-ART forms 246 prototypes (575 KB) but at only 67.4\% task accuracy. Our task mappers all build upon input from an ImageNet-pretrained feature extractor. The preferred feature embedding was the one to yield best performance in the fine-grained oracle-PSP-BD upper bound baseline: for cifar100 this was a Resnet-34 and for 8-dsets, Alexnet (see \cite{supplementary}).

\begin{table*}[t]
\centering
\caption{\textbf{Fine-Grained Classification Performance and Memory Usage for each Task-Mapper and Baseline Method.} Task(\%) is task mapping accuracy and Main(\%) fine-grained classification performance. The first set of results are task-dependent with PSP-BD backbones. The (ours) indicates a task-mapper we propose. The second set includes task-independent baselines. Best results in bold.}
\begin{tabular}{cccccccccc}
\toprule
Method &  \multicolumn{3}{|c|}{8-dsets} &  \multicolumn{3}{|c|}{Permuted-MNIST-25} &
\multicolumn{3}{|c|}{Cifar100} \\
\midrule
 Task-Dependent  & Memory(KB)   & Task(\%)    & Main(\%)   & Memory(KB)   & Task(\%)    & Main(\%) & Memory(KB)   & Task(\%)    & Main(\%)\\
\midrule
Oracle  &  166 & 100.0   & 58.0   & 399  & 100.0 & 91.2 & 207 &  100.0 & 76.9 \\
Random   &  166 & 13.0   & 10.9    & 399  & 3.9 & 13.3 & 207 &  10.9 & 26.0 \\
Entropy   &  166 & 27.1   & 25.2    & 399 &  78.6 & 76.9 & 207 &  34.3 & 44.9 \\
AE-gates   &  1,804 & 79.6   & 45.4    & 1,183 &  100.0 & 91.2 & 2,255 &  56.6 & 53.4 \\
KM-Heads$\dagger$   &  287 & 32.0   & 20.7  & 345 & 26.7 & 18.1 & 757 & 21.7 & 32.1 \\
KM-Heads (ours) &  453 & 3.1   & 12.5  & 499 & 63.9 & 66.6 & 227 & 20.9  & 32.5 \\
NMC (ours)  &  207 & 92.9   & 55.3     & \textbf{475}  & \textbf{100.0} & \textbf{91.2} & 309 &  66.8 & 59.1\\
GMMC (ours) & \textbf{248} & \textbf{93.9}   & \textbf{55.8} & \textbf{556}  & \textbf{100.0} & \textbf{91.2} & \textbf{411} &  \textbf{68.0} & \textbf{59.4} \\
ART  (ours)  &  741 & 67.4   & 36.5   & 4,181  & 93.2 & 83.5 & 2,033 & 34.1   & 44.6 \\
ARTMAP (ours)   &  1,304 & 62.5  & 35.8   & 5,567  & 83.6 & 79.5 & 9,554 & 45.8  & 47.3 \\
PCR (ours)   &  376 & 91.6  & 54.5   & 982  & 98.8 & 86.0 &709 &  57.3 & 54.6  \\
PCR-E (ours)   &  2,932 & 75.2   & 45.8  &  845 & 66.3   & 69.2  &  959 & 48.7   & 47.0\\
\midrule
Task-Independent & (KB)   & (\%)    & (\%)   & (KB)   & (\%)    & (\%) & (KB)   & (\%)    & (\%)\\
\midrule
EWC &  0 & -   & 15.2*    & 0  & - & 45.8* & 0&  - & 35.3** \\
Vanilla-Replay &  66,060 & -   & 19.3**    & 20,070  & - & 20.0** & 4,719 &  - & 31.3** \\
GEM &  66,060 & -    & 10.0*   &   20,070  & - & 79.3* &4,719 &  - &  32.5* \\
\bottomrule
\end{tabular}
\vspace{4pt}
\footnotesize{Task-Independent: best results were obtained by *using a shared output head; **using one node per class. $\dagger$ has no PSP-BD and is multi-head.}
\end{table*}

\begin{figure*}[t]
    \begin{center}
        \includegraphics[width=17.5cm]{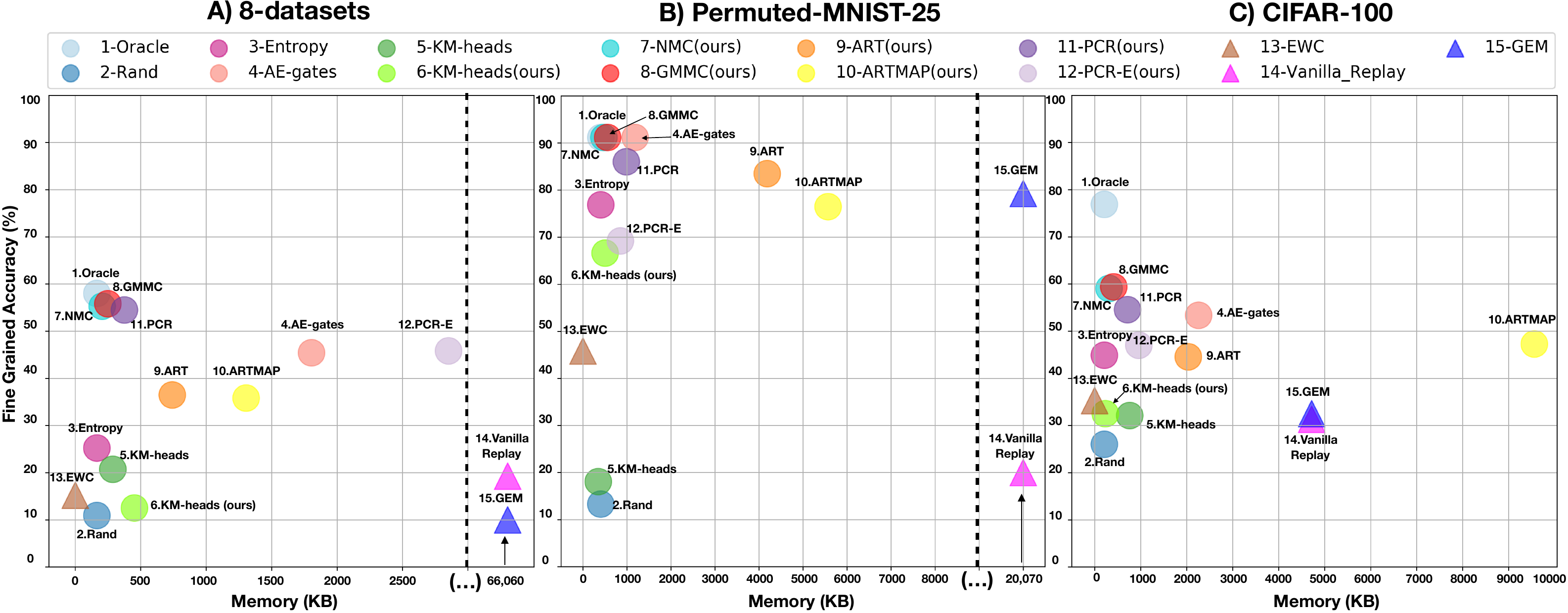}
		\caption{\textbf{Fine-grained classification performance versus model's memory usage.} Squares are task-dependent models and triangles, task-independent. Absolute upper-bound is Oracle + PSP-BD. Our full pipeline, with GMMC, NMC or PCR + PSP-BD classifier, is the closest to the optimal upper-left corner.}
        \label{fig:model}
    \end{center}
    \vspace{-10pt}
\end{figure*}

\subsection{Task-Dependent Vs Task-Independent Performances}

In our full pipeline, task-mapper + PSP-BD classifier, we keep only the best parameter configuration of each task-mapper according to the memory-performance tradeoff formula. Table I contains fine-grained classification performance of our task-mappers when combined with the fine-grained PSP-BD classifier. We compare our model variants with task-dependent (KM-Heads, AE-Gates and Entropy) and task-independent baselines (EWC, GEM and Vanilla-Replay) with respect to performance and memory storage usage. Memory-performance tradeoff for fine-grained classification is shown in Figure 6. The upper bound is the Oracle + PSP-BD. For full pipelines, memory is measured as space occupied by additional parameters needed to ensure remembering; this excludes the original weights of the backbone network. For instance, in oracle + PSP-BD the 160 KB memory usage refers exclusively to PSP and BD components.

Despite  the simplicity of our best mappers, GMMC, NMC, PCR, when combined with the strong PSP-BD classifier, they enable much better performance than the task-independent and task-dependent baselines. In figure 6, GMMC, NMC and PCR mappers cluster closely to the oracle baseline as well as the optimal upper-left corner. In contrast, all task-independent methods (EWC, GEM and Vanilla-Replay) show very poor memory usage and lower performance. From Figure 6, we also note how 8-dsets and Cifar100 are in general much harder benchmarks than Permuted-MNIST. In the latter, most models achieve above 70\% accuracy after 25 tasks, including GEM, KM-Heads and Entropy, which fall drastically to below 40\% and 20\% in Cifar-100 and 8-dsets, respectively. 

When comparing PCR and PCR-E, we observe that using features from a fixed-extractor as input to task mappers in general worked much better than feeding task-specific logit embeddings.
Similarly, the task-dependent baselines KM-heads and Entropy, which also use logit embeddings as input to a task classifier, also perform poorly. Additionally, AE-gates was shown to perform better than other task-dependent baselines (Entropy and KM-heads). Yet, in the harder benchmarks of 8-dsets and Cifar-100, AE-gates still significantly underperformed our best proposed task-mappers (GMMC, NMC and PCR) both in terms of accuracy and memory usage. For instance, when comparing AE-gates to GMMC in 8-dsets, the former obtains a task accuracy of 79.6\% at a cost of 1,804 KB whereas our GMMC model obtains 93.9\% task accuracy at only 248 KB. In Table II we compute the percentage increase in performance and  decrease in accuracy from our best task-mapper + PS-BD models relative to the oracle + PSP-BD upper bound.

\vspace{-2pt}
\subsection{Incremental Inter x Intra Dataset learning}
Our approach is shown to work particularly well for Inter-dataset incremental learning, i.e., 8-dsets and Permuted MNIST (Table II). Inter-dataset variability is in general larger than intra-dataset, making task mapping easier. Several factors contribute to this, one being greater diversity of low-level image statistics between different datasets.
Thus, even in a difficult benchmark like 8-dsets, we achieve very high task classification performances. For instance, in 8-dsets, while tasks such as VOC and ACTIONs are close both semantically and in low-level statistics, we also have much more distant tasks such as VOC with relation to SVHN. Meanwhile, in the Intra-dataset modality (Cifar100), while each superclass enforces some semantic grouping, i.e., fish versus flowers, the low-level statistics among tasks is very similar since all images are sampled from the same source. 
\vspace{-2pt}

\begin{table}[h]
\centering 
\caption{Best Task Mapper with respect to Oracle. Percentages taken from the best versions of our full model (GMMC/NMC + PSP-BD) with respect to upper-bound oracle + PSP-BD.} 
\renewcommand{\arraystretch}{1.3} 
\begin{tabular}{p{3cm}p{1.1cm}p{1.1cm}p{1.1cm}} 
\cmidrule(l){1-4} 
{} & 8-dsets & Permuted & Cifar100\\ 
\midrule 
Performance Decrease (\%) & 3.79 & 0.22  & 22.75 \\ 
Memory Increase (\%) & 0.03 & 5.37 &  0.09\\ 
\midrule 
\end{tabular}
\label{tab:template} 
\end{table}

\section{Conclusion}
We propose and compare several task mapping models that impose only very modest memory storage increments when coupled to a fine-grained classification model. 
We find that when using our best performing task-mapper with a state-of-the-art fine-grained classifier, we perform better than with baseline mappers and also much superior to task-independent methods both in terms of accuracy and memory usage.

Recent work in meta and adaptive learning suggests that a crucial element for generalizable learning is a good representation \cite{dhillon}. 
In fact, we found that for task mapping, inputs from a shared robust fixed extractor enabled much better performance then inputs from task-specific embeddings.
By exploring a good representation, we show that simple methods of task classification can work well both for incorporating new information and for not forgetting past information. 
A pivotal question for the field is then how to generate more generalizable visual embeddings, from which can arise simpler and more memory efficient continual learning algorithms.

\section*{Acknowledgment}
This work was supported by C-BRIC (part of JUMP, a Semiconductor Research Corporation (SRC) program sponsored by DARPA), the Intel Corporation, and CISCO.




%

\end{document}